\documentclass[conference]{IEEEtran}
\IEEEoverridecommandlockouts
\usepackage{cite}
\usepackage{amsmath,amssymb,amsfonts}
\usepackage{graphicx}
\usepackage[table]{xcolor}

\usepackage[ruled,vlined]{algorithm2e}

\SetKwInput{KwInput}{Input}   
\SetKwInput{KwOutput}{Output} 

\usepackage[utf8]{inputenc} 
\usepackage[T1]{fontenc} 
\usepackage{graphicx} 
\usepackage{booktabs} 
\usepackage{multirow}
\usepackage{array}

\usepackage{graphicx}
\usepackage{textcomp}
\usepackage{xcolor}
\def\BibTeX{{\rm B\kern-.05em{\sc i\kern-.025em b}\kern-.08em
    T\kern-.1667em\lower.7ex\hbox{E}\kern-.125emX}}
\begin{document}

\title{Beyond Augmentation: Empowering Model Robustness under Extreme Capture Environments\\
\thanks{}
}

\author{
	\IEEEauthorblockN{
		Yunpeng Gong\IEEEauthorrefmark{2}, 
		Yongjie Hou\IEEEauthorrefmark{1}, 
		Chuangliang Zhang\IEEEauthorrefmark{2}, 
		Min Jiang\IEEEauthorrefmark{2}\IEEEauthorrefmark{2}
	}
	
	\IEEEauthorblockA{\IEEEauthorrefmark{2}School of Informatics, Xiamen University, Xiamen, China}
	\IEEEauthorblockA{\IEEEauthorrefmark{1}School of Electronic Science and Engineering, Xiamen University, Xiamen, China}
	
	\IEEEauthorblockA{
		Email: fmonkey625@gmail.com, 
		23120231150268@stu.xmu.edu.cn, 
		31520231154325@stu.xmu.edu.cn, 
		minjiang@xmu.edu.cn
	}
	
	\thanks{\IEEEauthorrefmark{2}\IEEEauthorrefmark{2} Min Jiang is the corresponding author.}
}

\maketitle


\maketitle

\begin{abstract}
Person Re-identification (re-ID) in computer vision aims to recognize and track individuals across different cameras. While previous research has mainly focused on challenges like pose variations and lighting changes, the impact of extreme capture conditions is often not adequately addressed. These extreme conditions, including varied lighting, camera styles, angles, and image distortions, can significantly affect data distribution and re-ID accuracy.

Current research typically improves model generalization under normal shooting conditions through data augmentation techniques such as adjusting brightness and contrast. However, these methods pay less attention to the robustness of models under extreme shooting conditions. To tackle this, we propose a multi-mode synchronization learning (MMSL) strategy . This approach involves dividing images into grids, randomly selecting grid blocks, and applying data augmentation methods like contrast and brightness adjustments. This process introduces diverse transformations without altering the original image structure, helping the model adapt to extreme variations. This method improves the model's generalization under extreme conditions and enables learning diverse features, thus better addressing the challenges in re-ID. Extensive experiments on a simulated test set under extreme conditions have demonstrated the effectiveness of our method. This approach is crucial for enhancing model robustness and adaptability in real-world scenarios, supporting the future development of person re-identification technology.
\end{abstract}


\section{Introduction}

With the rapid development of computer vision and deep learning technologies, person-centric wide-area surveillance systems play a crucial role in public safety and industrial monitoring\cite{5,6,3,4}. Wide-area surveillance involves monitoring and analyzing large-scale environments through different camera setups, providing essential information for decision-making and security. However, the models in these application scenarios face increased demands for robustness due to extreme shooting conditions.

In industrial settings, especially in construction sites, mining areas, and manufacturing plants, challenges such as dust or smoke, extreme lighting, temperature variations, vibrations, and pollution pose severe obstacles for machine vision systems~\cite{1,2,7,8}. These conditions demand special requirements for the accuracy and robustness of visual models, particularly in tasks like quality control and safety monitoring, where precise identification of target objects under diverse lighting and background conditions is crucial to avoid potential major accidents. Consequently, researchers need to develop algorithms that can adapt to these challenging conditions and train and test these models with real-world industrial data to ensure reliability under various circumstances.

In applications like urban surveillance, traffic monitoring, and border security, where different camera settings, complex lighting, and extensive scene coverage are common, models are susceptible to color domain deviations. Additionally, surveillance systems often operate under extreme conditions such as adverse weather, low lighting, or long-distance shooting. The impact of these factors often leads to unstable model performance, as they impose high demands on the visual robustness of the models. Therefore, the key research question becomes how to enhance the performance of these surveillance systems under various extreme conditions using advanced image processing techniques and deep learning.

\begin{figure*}[htbp]
	\centerline{\includegraphics[width=0.9\textwidth]{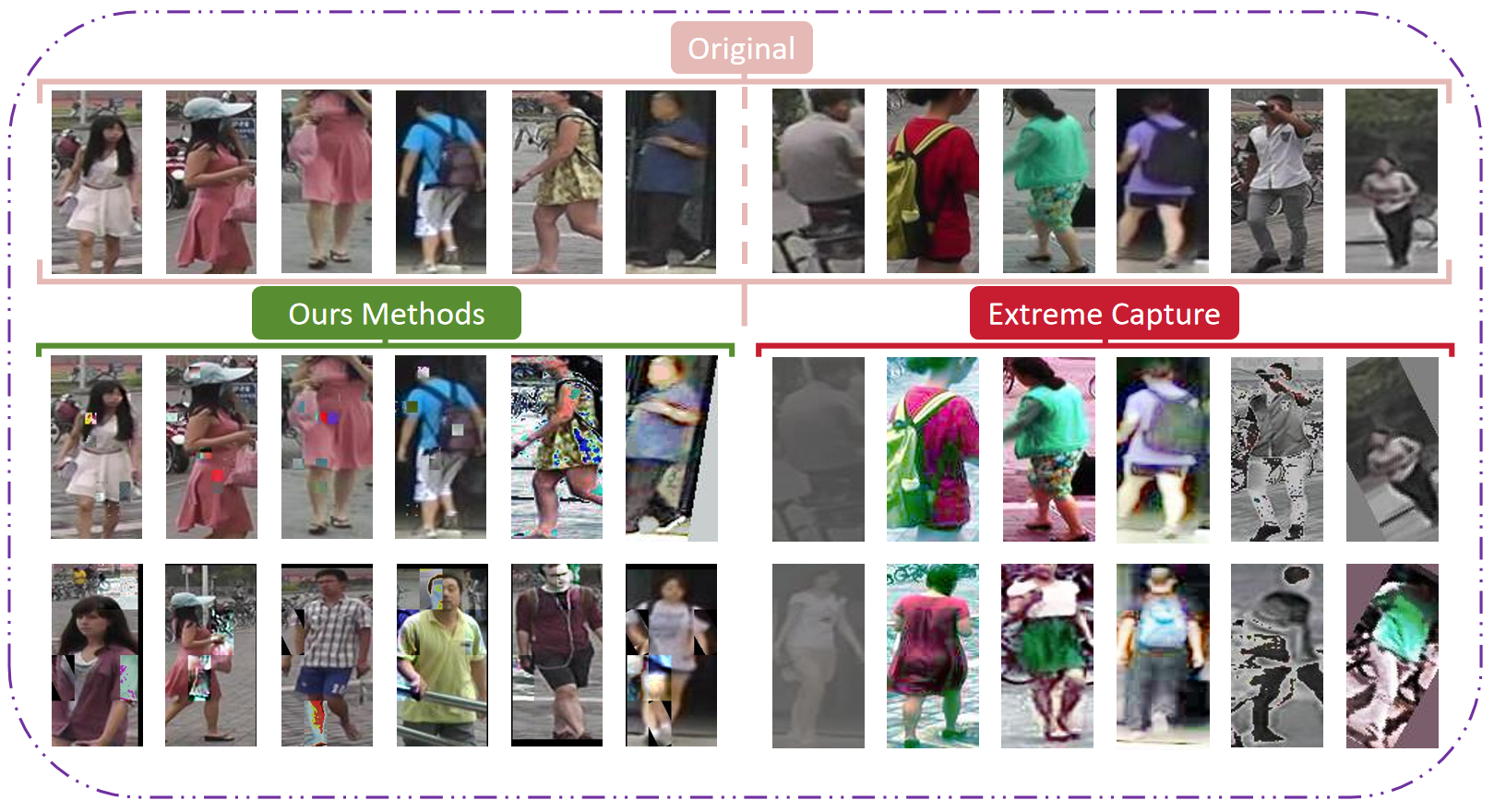}}
	\caption{
		The first row in the image consists of normal capture images. In 'Extreme Capture', we present example images simulating extreme shooting conditions. These include simulations of low lighting or heavy fog, camera style shifts, overexposure, data corruption, and camera position faults, among other extreme factors. In 'Our Methods', schematic representations of the proposed approach are provided. This approach progressively learns various extreme shooting conditions by randomly applying different augmentation methods to different image grids.}
	\label{fig}
\end{figure*}

Currently, to improve model performance under different conditions, researchers typically employ various data augmentation~\cite{7,1,2} techniques such as adjusting brightness, contrast, and adding noise to simulate extreme conditions. This approach aims to train models to better generalize to various input conditions. In some cases, research focuses on specific types of extreme conditions, such as nighttime or image recognition under adverse weather conditions. These studies often involve developing or adjusting models to adapt specifically to these particular contexts. In fields like autonomous vehicles and drone navigation, models need to maintain accuracy and reliability in various extreme environments. Therefore, research in these application areas poses special requirements for the robustness of models under actual physical conditions.

Person Re-identification (re-ID)~\cite{3,5,6,2,12,13,32,33,34,35}, as a crucial security monitoring technology, faces the aforementioned challenges. Therefore, this paper focuses on studying the robustness of models under extreme shooting conditions with re-ID as the target. Current research in re-ID often improves model generalization under normal shooting conditions through data augmentation techniques like adjusting brightness and contrast. However, these methods pay less attention to the robustness of models under extreme shooting conditions~\cite{1,2}. To address this issue, we propose a multi-mode synchronization learning strategy based on the idea of data augmentation~\cite{2}. This method involves dividing images into a grid, randomly selecting grid blocks, and applying data augmentation methods such as contrast and brightness adjustments. This process introduces diverse transformations without altering the original image structure, helping the model adapt to extreme variations. The proposed method enhances the model's generalization under extreme conditions, enabling it to learn diverse features to better address the challenges in re-ID. Extensive experiments conducted under simulated extreme conditions have demonstrated the effectiveness of our approach. This method is crucial for enhancing the robustness and adaptability of models in real-world scenarios, supporting the future development of person re-identification technology.

\begin{figure*}[htbp]
	\centerline{\includegraphics[width=1\textwidth]{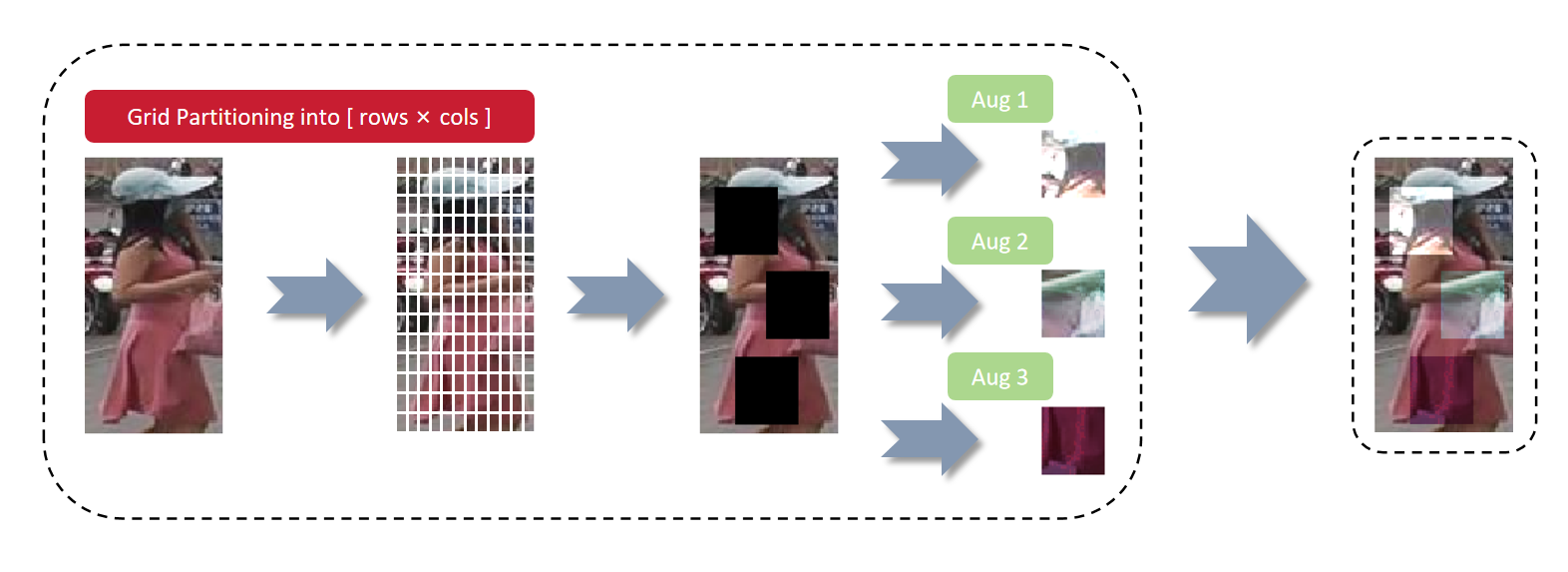}}
	\caption{
Diagram illustrating the Multi-Mode Synchronization Learning (MMSL) strategy. Initially, partition the images in the training dataset into a grid of $rows$ $\times$ $cols$. Next, randomly select a subset of tiles. For these chosen image tiles, randomly pick the corresponding number of data augmentations from the AutoAugment library and apply them to the selected image regions.}
	\label{fig}
\end{figure*}

The main contributions of this paper are summarized as follows:

$\bullet$ We propose a novel learning strategy termed Multi-Mode Synchronization Learning (MMSL) to enhance the robustness of person re-identification (re-ID) models under extreme shooting conditions. This strategy involves partitioning images into grids, randomly selecting grid blocks, and applying data augmentation techniques such as contrast and brightness adjustments. It introduces diverse transformations without altering the original image structure, aiding the model in adapting to extreme variations.

$\bullet$ Addressing the challenges posed by extreme shooting conditions in wide-area surveillance systems, our proposed strategy contributes to improving the robustness and adaptability of models in real-world scenarios. Wide-area surveillance plays a crucial role in industrial production and safety monitoring, and extreme shooting conditions can lead to model performance instability. Our method provides an effective approach to mitigate this issue.

$\bullet$ We conduct extensive experiments to validate the effectiveness of the proposed strategy. Particularly in simulated extreme conditions, our method demonstrates improved model generalization, enabling the learning of diverse features to better address the challenges in re-ID tasks. This empirical evidence supports the practical applicability of our approach.

\section{Related Work}
In the context of varied visual scenarios, the intrinsic complexity of tasks introduces a comparable complexity in their data requirements. Failing to adequately address these intricate data needs can result in issues such as model overfitting and insufficient generalization to the training data. Enhancing generalization capabilities has consistently remained a central focus of convolutional neural network (CNN) research, and data augmentation has demonstrated notable effectiveness in enhancing the model's generalization ability.

\subsubsection{Person Re-identification}
Person Re-identification (re-ID) stands as a pivotal task in computer vision, focusing on the recognition and tracking of individuals across different time points and camera views within video sequences or images. This task holds significant applications in surveillance, video analysis, and intelligent transportation systems. However, re-ID faces formidable challenges, encompassing pose variations, changes in lighting conditions, occlusions, and low resolutions~\cite{23}. The subtle discrepancies in pedestrian appearances further compound the difficulty of distinguishing between distinct individuals. Addressing the intricacies of re-ID necessitates the extraction of robust features from images or video frames.

Deep learning, particularly CNNs, is widely employed to learn high-level features from images within the domain of re-ID. Feature extraction networks tailored for this task often integrate pre-trained CNN architectures, such as ResNet~\cite{24} and PCB. Metric learning is a key technology in re-ID, crucial for quantifying the similarity between two pedestrian images. Commonly used metric learning methods include Euclidean distance and cosine similarity. Learning a suitable metric ensures that the feature representations of the same individual are closely aligned, while those of different individuals are distinctly separated. This emphasis on metric learning contributes to the efficacy of re-ID systems.

The importance of re-ID extends beyond mere technological advancements, finding practical applications in surveillance, video analytics, and the optimization of intelligent transportation systems. By overcoming challenges related to pose variations, lighting changes, occlusions, and low resolutions, robust feature extraction through deep learning methodologies, and effective metric learning techniques, re-ID systems play a critical role in enhancing the security and efficiency of various real-world scenarios.

\subsubsection{Data Augmentation}
Data augmentation is a widely employed technique, especially in the field of computer vision. It enhances the diversity of a dataset by applying a series of transformations to the original data. This not only aids the model in learning more robust and generalized features but also contributes to improving model performance on limited datasets.

Commonly utilized data augmentation techniques include: Random Cropping~\cite{14}, by randomly cropping a part of an image, the model focuses on different regions, enhancing its ability to recognize local features. This is particularly beneficial for tasks like object detection and classification. Horizontal Flipping and Rotation~\cite{15}, flipping images horizontally or vertically and rotating them helps the model learn orientation invariance, crucial for understanding the direction and shape of objects within images. Scale Transformation, scaling images aids the model in recognizing objects of various sizes, especially important in tasks like object detection. Noise Injection~\cite{20}, adding random noise to images improves the model's robustness, enabling it to handle imperfections in real-world images more effectively. Geometric Transformations~\cite{22}, introducing geometric transformations, such as skewing or distorting, allows the model to learn more complex geometric deformations. Shearing, moving a part of an image through shearing creates new perspectives and compositions, assisting the model in understanding different combinations of objects. CutMix~\cite{16}, an advanced data augmentation method, goes beyond simple cropping and pasting of image blocks; it replaces a region of one image with that of another. Such mixing aids the model in better understanding relationships between different regions. Introducing noise blocks or randomly deleting pixels in the image helps regularize the network, preventing the model from overly relying on specific parts and addressing occlusion issues.

The application of these data augmentation techniques enhances the dataset's variability, leading to a more robust and versatile model capable of handling a wide range of scenarios in computer vision tasks. In enhancing the model's robustness under extreme shooting conditions, Color Jittering~\cite{21} involves altering the image's brightness, contrast, and saturation to better adapt the model to varying lighting and color conditions.  Random Erasing~\cite{17} achieves regularization by introducing noise blocks or randomly deleting pixels in the image, preventing the model from excessive dependence on specific regions and aiding in overcoming occlusion challenges. AutoAugment~\cite{18} is an automated data augmentation strategy that integrates a series of augmentation methods. By searching for suitable augmentation strategies, such as image cropping transformations, color adjustments, and brightness/contrast adjustments, it aims to enhance model performance. It is widely applied to address issues related to significant dataset domain disparities. AugMix~\cite{19} aims to enhance the performance and robustness of deep learning models by introducing diversity and complexity. The method involves applying multiple data augmentation operations to the input image and blending these distorted images in a mixed form, creating diverse versions of the original training images. Please note that AugMix involves the blending of images from different samples, disrupting the original structure of the images. Consequently, as demonstrated in our subsequent experiments, its performance under extreme shooting conditions is less than ideal. In contrast, our proposed method avoids compromising the structural integrity of the original images. Simultaneously, it enables the model to learn various extreme variations, enhancing the robustness of the model to extreme shooting conditions. In future work, we will also consider integrating evolutionary computation techniques~\cite{27,28,29,30,31} to extend our method for adaptive optimization.

\begin{algorithm}[t]
	\DontPrintSemicolon
	\caption{Multi-Mode Synchronization Learning Strategy}
	
	\KwInput{
		
		\\\hspace*{3em}Image $i$,\\ \hspace*{3em}Grid Size: $({rows}, {cols})$,\\
		\hspace*{3em}Max Number of Tiles $n$,\\
		\hspace*{3em}Total Probability $p_{t}$,\\ \hspace*{3em}Global Probability $p_{g}$.}
	
	\KwOutput{\\\hspace*{3em}Augmentation Image \( I \).}
	
	\textbf{Initialization}: \\\hspace*{3em}\( p_g \leftarrow \text{Rand}(0, 1) \),\( p \leftarrow \text{Rand}(0, 1) \).\\
	\If{\( p_g \leq p \)}{
		\hspace*{1em} $\bullet$ Randomly select a data augmentation method from the augmentation library and apply it to \( I \).\\
		\Return \( I \).
	}
	\ElseIf{\( p_t < p \)}{
		\( N \leftarrow \text{Rand}(0, n), \)\\
		\For{\( num \) in \( \text{range}(0, N) \)}{
			$\bullet$ Randomly pick a grid from the set of grids $P_i$ with [$rows$ $\times$ $cols$].\\
			$\bullet$ Randomly select a data augmentation method from the augmentation library and apply it to the grid $P_i$.\\
		}
		\Return \( I \).
	}
\end{algorithm}

\section{Proposed Method}
Drawing inspiration from AutoAugment~\cite{18}, this study introduces a novel augmentation strategy called multi-mode synchronization learning (MMSL). In comparison to its predecessor, it enables the model to learn and adapt to extreme variations in data distribution more efficiently. This approach enhances the model's generalization ability under extreme conditions, allowing it to capture invariant features in data with drastic variations and thereby improving robustness to visual challenges.

Our MMSL strategy consists of two components, namely global differentiation learning and multi-grid differentiation learning. global differentiation learning is a specific case of multi-grid differentiation learning.

\subsubsection{Global differentiation learning}
AutoAugment integrates a series of data augmentation operations, including ShearX and ShearY for horizontal or vertical shear transformations, introducing variations in the shapes of objects within images. This aids the model in learning to handle images from different angles and perspectives, thereby improving the model's robustness. TranslateX and TranslateY perform horizontal or vertical translation transformations, causing the movement of the image's object positions. This enhances the model's adaptability to changes in object positions and improves robustness to spatial transformations. Additionally, rotating the image by a certain angle helps the model adapt to rotational transformations, addressing issues related to rotational invariance.

In addition to these operations, AutoAugment includes color-related operations such as Color, Posterize, Solarize, Contrast, Sharpness, and Brightness. These operations enrich the color features of images, strengthening the model's resilience to changes in lighting conditions and color. Introducing AutoContrast, Equalize, and Invert further achieves automatic adjustments to contrast, histogram equalization, and color inversion, helping enhance the recognizability of images. The overall goal of these data augmentation operations is to introduce diversity into the training data, thereby enhancing the model's generalization ability. This enables the model to better adapt to different visual scenarios, lighting conditions, angles, and positional transformations, ultimately improving performance in practical applications. Furthermore, the enhanced diversity of data aids in mitigating the model's sensitivity to overfitting.

In the data loading, it randomly samples $Q$ images of per person and $M$ identities to constitute a training batch, which size is $B=Q\times M$. The set is denoted as $I=\{I_k|k=1,2,...,Q\times M\}$.

The proposed method randomly performs global transformation on the training batch with a probability, and then inputs the processed images into the model for training. This transformation process can be defined as:
\begin{equation}
	I'_k = t(I_k)
\end{equation}
where $t(\bullet)$ represents the augmentation function, which randomly selects one augmentation method from the AutoAugment library to transform the image. $y_k$ is the label of the sample, the label of converted image remains unchanged, so
\begin{equation}
	(I_k|y_k) = (I'_k|y_k)
\end{equation}
\subsubsection{multi-grid differentiation learning}
In this methodology, we divide the images in the training dataset into a grid of $rows$ $\times$ $cols$. Subsequently, a random number $N$ is generated from the range of the maximum number of tiles $n$, determining how many grid regions in the image will undergo transformations. For these selected $N$ image tiles, we randomly extract the corresponding number of data augmentations from the AutoAugment library and apply them to the chosen image regions. The entire process can be expressed through the following equations:

\begin{equation}
	P = RandPatch(Grid(I_k,rows,cols),n),
\end{equation}
Here, Grid($I_k$, $rows$, $cols$) partitions the image $I_k$ into a grid of $rows$ $\times$ $cols$. Additionally, RandPatch($\bullet$, $n$) randomly selects $n$ blocks from all image grid blocks for image transformations.
\begin{equation}
	P'_i = RandAugment(I_k(P_i)) (P'=[P'_i|i=1,2,...,n]),
\end{equation}
random data augmentation is applied to the i-th selected image block $P_i$ from the image $I_k$, where the augmentation methods are drawn from the AutoAugment data augmentation library. The transformed image block is denoted as $P'_i$. The overall image transformation process for our MMSL strategy can be expressed as follows:
\begin{equation}
	I'_k = I_k- P + P' .
\end{equation}
$y_k$ is the label of the sample, the label of converted image remains unchanged, so
\begin{equation}
	(I_k|y_k) = (I'_k|y_k)
\end{equation}

During the model training phase, our Multi-Mode Synchronization Learning (MMSL) strategy is stochastically applied to the training batch, introducing diverse modes within the same image while preserving the structural integrity of objects.

\begin{figure*}[htbp]
	\centerline{\includegraphics[width=1\textwidth]{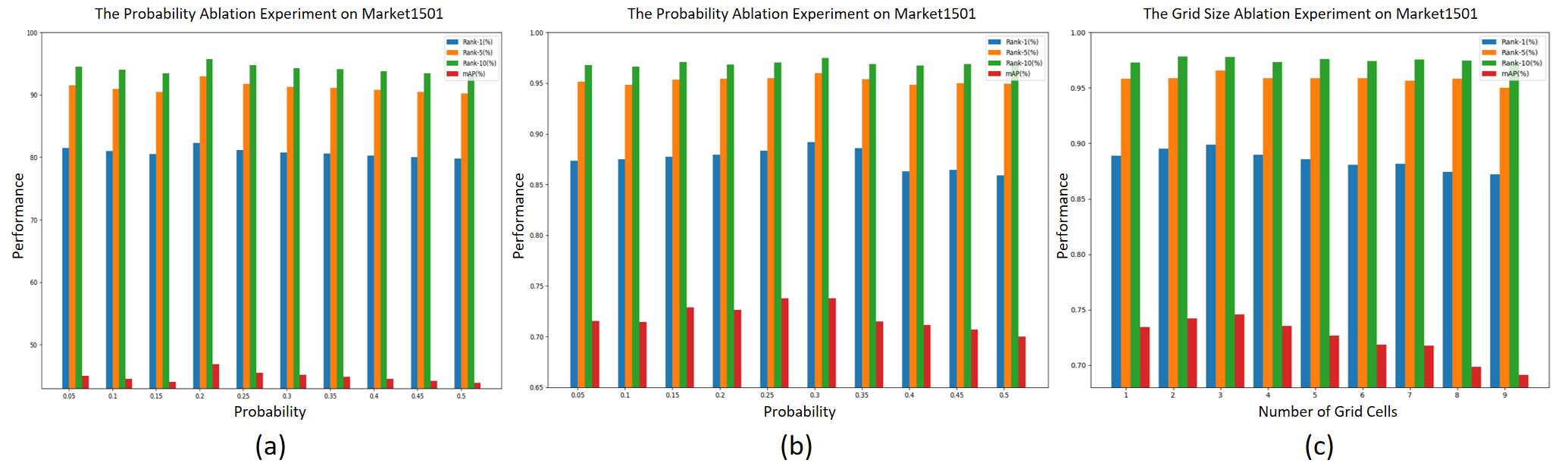}}
	\caption{
		Multi-Mode Synchronization Learning (MMSL) strategy ablation study. (a) Experiment on setting the probability of global augmentation. (b) Experiment on setting the probability of local augmentation. (c) Experiment on the ratio of local augmentation blocks in a $3\times3$ grid.}
	\label{ab}
\end{figure*}

\begin{figure*}[htbp]
	\centerline{\includegraphics[width=1\textwidth]{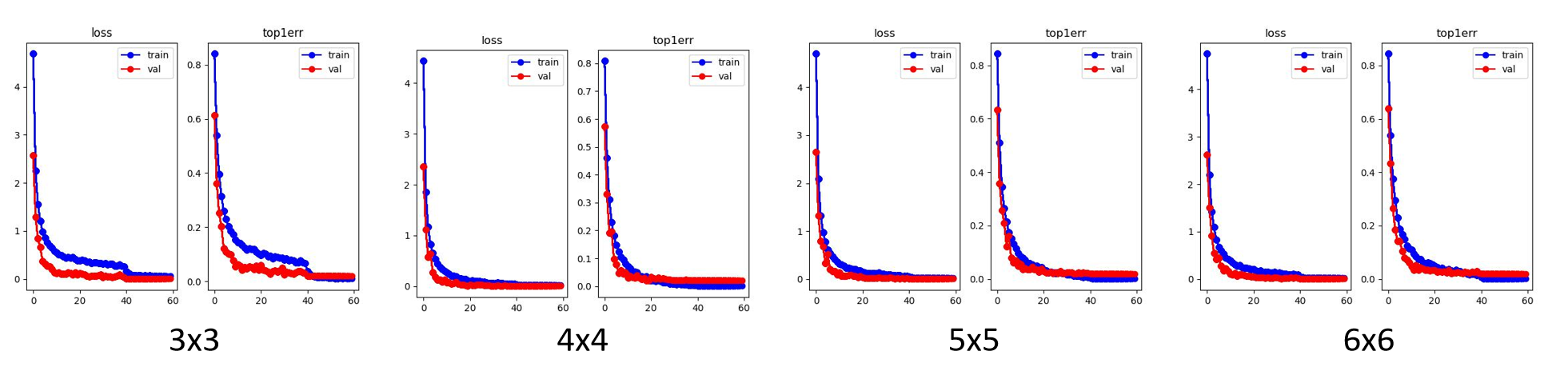}}
	\caption{
		Grid Size Ablation Study: Our Multi-Mode Synchronization Learning (MMSL) strategy showcases training curve graphs when training the model with different grid sizes, illustrating training loss and error rates of top-1 retrieval results.}
	\label{nxn}
\end{figure*}

\begin{table*}[]
	\caption{Performance comparison of different methods on the original Market1501 dataset and the simulated extreme capture conditions Market1501 dataset. Rank at \( r \) accuracy (\%) and mAP (\%) are reported.}
	\centering
	\vspace{-5pt} 
	\renewcommand{\arraystretch}{1}
	\resizebox{0.95\textwidth}{!}{%
		\begin{tabular}{|l@{\hspace{6pt}}|l@{\hspace{6pt}}|c@{\hspace{6pt}}|c@{\hspace{6pt}}|c@{\hspace{6pt}}|c@{\hspace{6pt}}|c@{\hspace{6pt}}|c@{\hspace{6pt}}|c@{\hspace{6pt}}|c@{\hspace{6pt}}|}
			\hline
			\rowcolor{gray!50}
			\multicolumn{2}{|c|}{{Settings}} & \multicolumn{4}{c|}{{Original}} & \multicolumn{4}{c|}{{Extreme Capture}} \\
			\cline{3-10}
			\rowcolor{gray!25}
			Method & Venue & \( r = 1 \) & \( r = 5 \) & \( r = 10 \) & mAP & \( r = 1 \) & \( r = 5 \) & \( r = 10 \) & mAP \\
			\hline
			baseline  & TMM  & 88.84 & 95.36 & 97.15 & 71.59 & 65.79 & 86.04 & 84.10 & 40.74\\
			AugMix  &ICLR   & 86.10 & 95.07 & 96.97 & 67.12 & 68.73 & 88.37 & 86.96 & 42.09 \\
			AutoAugment  &CVPR   & 85.89 & 94.41 & 96.46 & 66.44 & 82.30 & 93.02 & 95.75 & 46.84 \\
			Our Method  &{---------}  & 89.19 & 95.99 & 97.47 & 73.80 & 81.50 & 92.96 & 95.69 & 45.52 \\
			\hline
		\end{tabular}%
		
	}
	\label{market}
\end{table*}

\begin{table*}[]
	\caption{Performance comparison of different methods on the original Duke dataset and the simulated extreme capture conditions Duke dataset. Rank at \( r \) accuracy (\%) and mAP (\%) are reported.}
	\centering
	\vspace{-5pt} 
	\renewcommand{\arraystretch}{1}
	\resizebox{0.95\textwidth}{!}{%
		\begin{tabular}{|l@{\hspace{6pt}}|l@{\hspace{6pt}}|c@{\hspace{6pt}}|c@{\hspace{6pt}}|c@{\hspace{6pt}}|c@{\hspace{6pt}}|c@{\hspace{6pt}}|c@{\hspace{6pt}}|c@{\hspace{6pt}}|c@{\hspace{6pt}}|}
			\hline
			\rowcolor{gray!50}
			\multicolumn{2}{|c|}{{Settings}} & \multicolumn{4}{c|}{{Original}} & \multicolumn{4}{c|}{{Extreme Capture}} \\
			\cline{3-10}
			\rowcolor{gray!25}
			Method & Venue & \( r = 1 \) & \( r = 5 \) & \( r = 10 \) & mAP & \( r = 1 \) & \( r = 5 \) & \( r = 10 \) & mAP \\
			\hline
			baseline  & TMM  & 79.48 & 89.04 & 92.10 & 61.96 & 73.74 & 85.68 & 89.00 & 35.36 \\
			AugMix  &ICLR   & 73.24 & 84.64 & 88.24 & 52.42 & 70.57 & 82.35 & 86.55 & 42.47 \\
			AutoAugment  &CVPR   & 74.14 & 85.09 & 88.06 & 53.04 & 71.40 & 83.30 & 87.07 & 47.93 \\
			Our Method  & {---------}   & 79.26 & 88.78 & 91.92 & 61.57 & 71.45 & 84.42 & 88.46 & 49.21 \\
			\hline
		\end{tabular}%
		
	}
	\label{duke}
\end{table*}

	\begin{table}[!htbp]
		\caption{Performance comparison of different methods in cross-domain testing between the original Market1501 and original Duke datasets.}
		\centering
		\renewcommand{\arraystretch}{1.2}
		\resizebox{\columnwidth}{!}{%
			\begin{tabular}{|l@{\hspace{6pt}}|l@{\hspace{6pt}}|c@{\hspace{6pt}}|c@{\hspace{6pt}}|c@{\hspace{6pt}}|c@{\hspace{6pt}}|}
				\hline
				\rowcolor{gray!50}
				\multicolumn{2}{|c|}{{Settings}} & \multicolumn{4}{c|}{{Original}} \\
				\cline{3-6}
				\rowcolor{gray!25}
				Method & State & \( r = 1 \) & \( r = 5 \) & \( r = 10 \) & mAP \\
				\hline
				\multirow{4}{*}{Market->Duke}  & baseline & 33.43 & 48.11 & 54.66 &17.53 \\
				& AugMix  & 30.87 & 46.09 & 53.59 & 15.36 \\
				& AutoAugment  & 38.06 & 54.62 & 60.86 & 19.08 \\
				& Ours   & 41.2 & 57.49 & 64.13 & 22.34 \\
				\hline
				\multirow{4}{*}{Duke->Market}  & baseline & 43.58 & 61.43 & 68.61 & 18.21 \\
				& AugMix  & 40.10 & 58.82 & 65.78 &14.43 \\
				& AutoAugment  & 44.19 & 62.42 & 70.04 & 17.41 \\
				& Ours   & 44.21 & 62.52 & 70.13 & 17.89 \\
				\hline
			\end{tabular}%
		}
		\label{ori}
	\end{table}

\section{Experimental Analysis}
This section will demonstrate the effectiveness of the proposed method through a series of qualitative, comparative experiments, and black-box attack experiments.
\subsection{Datasets and Evaluation Criteria}
The proposed method undergoes evaluation on two person re-identification (ReID) datasets: Market-1501 \cite{9} and DukeMTMC \cite{10}. These datasets are widely acknowledged as the most representative and extensively employed in ReID research. The Market-1501 dataset comprises 12,936 images with 751 identities for training, 19,732 images with 750 identities, and 3,368 query images for testing. DukeMTMC-reID includes 16,522 training images of 702 identities, 2,228 query images of the other 702 identities, and 17,661 gallery images.

Consistent with prior research~\cite{9}, the evaluation utilizes Rank-k precision, Cumulative Matching Characteristics (CMC), and mean Average Precision (mAP) as standard metrics. Rank-1 precision represents the average accuracy of the top-ranked result corresponding to each cross-modality query image. mAP signifies the mean average accuracy, calculated by sorting query results based on similarity. The closer the correct result is to the top of the list, the higher the precision. These metrics collectively offer a comprehensive assessment of the proposed method's performance in comparison to existing works.

To simulate extreme shooting conditions and showcase the superiority of our proposed method over existing approaches, we directly apply AutoAugment to transform the gallery set of the test set in the original dataset. It is worth noting that, intuitively, evaluating performance on a dataset configured in this way would obviously favor models trained with AutoAugment. However, subsequent experiments demonstrate that even in such an unfair test, our proposed method still outperforms AutoAugment. This substantiates that our method can effectively enhance existing data augmentation techniques, exhibiting superior gains.

\subsection{Parameter Configuration and Ablation Experiments}
There are several parameters to be determined in our strategy, including the probability $p_1$ of global augmentation, the size of the grid, the number of blocks $k$ for local transformations, and the probability $p_2$ of local augmentation. In our experiments, we use ~\cite{25} as the baseline.

In order to determine at which ratio models trained under different augmentations exhibit better robustness in extreme shooting conditions, we experimented with various augmentation proportions. Regarding the probability of global augmentation, we conducted quantitative analysis experiments. We performed experiments with $p_g$ set to 5\%, 10\%, 15\%, ..., 50\%. The experimental results are shown in Fig.~\ref{ab}(a). From the graph, it can be observed that the performance is optimal when the probability of global augmentation is $probability=0.2$. Therefore, unless otherwise specified, we default to using $p_g=0.2$ in our subsequent experiments.

Regarding the settings for the grid size and the number of local enhancement blocks, we conducted quantitative analysis experiments. We performed experiments with grid sizes $2\times2$, $3\times3$, ..., $6\times6$ to quantitatively analyze the optimal ratio of local enhancement blocks to the entire image for each corresponding grid size. We determined the optimal number of blocks for the respective sizes by analyzing quantitatively for the cases of $3\times3$. The experimental results are shown in Fig.\ref{nxn}. From the graph, it can be observed that the performance is optimal when the grid size is $5\times5$. Additionally, from Fig.\ref{ab}(c), it can be observed that the performance is optimal when the optimal ratio of local enhancement blocks to the entire image is $1/3$ (The number of grid cells is 3). Hence, unless stated otherwise, we adopt the use of the corresponding number of blocks with a ratio of $1/3$ as the default configuration in our subsequent experiments.

When the optimal parameters for the global augmentation probability $p_g$ and the number of blocks for local transformations are determined, we fix the above parameters to evaluate the probability of local transformations. As shown in Fig.~\ref{ab}(b), the model achieves optimal performance when the probability of local transformations is $0.3$. Therefore, unless otherwise specified, we default to using $p_t=p_g+0.3$ in our subsequent experiments.

\subsection{Comparison experiment}

We compared our method with two relevant approaches, AutoAugment~\cite{18} and AugMix~\cite{19}. As shown in Tab.~\ref{market}, on the Market1501 dataset, our proposed method not only does not degrade the model's performance on the original dataset but also improves it, while other methods affect the model's performance to varying degrees. In 'Extreme Capture,' it can be observed that although extreme testing is simulated using AutoAugment, AutoAugment does not have a significant advantage in this test. Additionally, from Tab.~\ref{duke}, consistent performance is observed on the Duke dataset.

In addition to the aforementioned tests, we conducted cross-domain testing to better evaluate the model's generalization performance. Cross-domain person reidentification aims at adapting the model trained on a labeled source domain dataset to another target domain dataset without any annotation. It is pointed out by~\cite{26} that the higher accuracy of the model does not mean that
it has better generalization capacity. In response to the above potential problems, we use cross-domain tests to verify the robustness of the model. 'Market->Duke' indicates training on the Market dataset and testing on the Duke dataset, while 'Duke->Market' represents the reverse scenario. As observed in Tab.~\ref{ori}, in the 'Market->Duke' scenario, our proposed method outperforms AutoAugment by 3.26\% in mAP. In the 'Duke->Market' scenario, our proposed method surpasses AutoAugment by 0.48\% in mAP. These experimental results indicate that the proposed method is more beneficial for improving the model's generalization performance. Similar experiments were conducted on the simulated extreme test set, and the results demonstrated a high level of consistency, as illustrated in Tab.~\ref{extre}.

Through the aforementioned series of experiments, it is evident that augmenting only local regions in an image surpasses augmenting the entire image. Our proposed approach represents a superior implementation of existing data augmentation methods. It embodies an optimized form of augmentation, demonstrating outstanding performance.

\section{Conclusion}
In this paper, we proposed a novel approach, Multi-Mode Synchronization Learning (MMSL) strategy, to enhance the robustness of person re-identification (re-ID) models under extreme shooting conditions. Unlike conventional data augmentation methods that focus on normal shooting conditions, our strategy introduces diverse transformations to adapt models to extreme variations in data distribution. The MMSL strategy incorporates two key components, namely Global Differentiation Learning and Multi-Grid Differentiation Learning.

In the Global Differentiation Learning, we drew inspiration from AutoAugment to perform random global transformations on the training batch, addressing challenges such as shear, translation, rotation, and color variations. This component contributes to the model's ability to generalize under varying perspectives, angles, and lighting conditions, ultimately improving its robustness.

The Multi-Grid Differentiation Learning component involved dividing images into a grid, randomly selecting grid blocks, and applying data augmentation methods. This method introduces diverse transformations without altering the original image structure, aiding the model in adapting to extreme variations in data distribution. Through extensive experiments under simulated extreme conditions, we demonstrated the effectiveness of our approach in improving model generalization and addressing challenges in re-ID tasks.
\begin{table}[t]
	\caption{Performance comparison of different methods in cross-domain testing between the Market1501 and Duke datasets under simulated extreme capture conditions.}
	\centering
	\renewcommand{\arraystretch}{1.2}
	\resizebox{\columnwidth}{!}{%
		\begin{tabular}{|l@{\hspace{6pt}}|l@{\hspace{6pt}}|c@{\hspace{6pt}}|c@{\hspace{6pt}}|c@{\hspace{6pt}}|c@{\hspace{6pt}}|}
			\hline
			\rowcolor{gray!50}
			\multicolumn{2}{|c|}{{Settings}} & \multicolumn{4}{c|}{{Extreme Capture}} \\
			\cline{3-6}
			\rowcolor{gray!25}
			Method & State & \( r = 1 \) & \( r = 5 \) & \( r = 10 \) & mAP \\
			\hline
			\multirow{4}{*}{Market->Duke}  & baseline & 17.41 & 28.86 & 34.73 & 4.98 \\
			& AugMix  & 22.30 & 37.20 & 43.62 & 7.62 \\
			& AutoAugment  & 31.53 & 49.50 & 56.77 & 14.01 \\
			& Ours   & 31.50 & 49.32 & 56.06 & 13.83 \\
			\hline
			\multirow{4}{*}{Duke->Market}  & baseline & 22.52 & 36.93 & 45.82 & 6.57 \\
			& AugMix   & 26.75 & 44.71 & 52.93 &8.21 \\
			& AutoAugment  & 33.58 & 53.27 & 62.87 & 11.15 \\
			& Ours  & 33.70 & 53.94 & 63.03 &11.87 \\
			\hline
		\end{tabular}%
	}
	\label{extre}
\end{table}
Our proposed MMSL strategy contributes to the field of person re-identification by providing an effective method to enhance model robustness and adaptability in real-world scenarios. The ability to handle extreme shooting conditions is crucial for the practical deployment of re-ID models in wide-area surveillance systems, industrial monitoring, and security applications. The empirical evidence from our experiments supports the practical applicability of the proposed approach, showcasing its potential to significantly improve the performance of person re-identification models in challenging environments.

In future work, we plan to further explore the integration of additional data augmentation techniques and investigate the adaptability of the MMSL strategy to different fields. Additionally, we aim to conduct experiments with real-world data under extreme conditions to validate the robustness and practical utility of our approach in diverse application scenarios. The ongoing development of person re-identification technology demands innovative strategies to address emerging challenges, and we believe that the proposed MMSL strategy represents a step forward in this direction.

\section*{Acknowledgment}

This work was partly supported by the National Natural Science Foundation of China under Grant No. 62276222

\vspace{12pt}
\color{red}

\end{document}